\documentclass[10pt]{article}
\usepackage[left=2.2cm,right=2.2cm,top=2.5cm,bottom=2.5cm]{geometry}

\usepackage{booktabs}
\usepackage{amsmath}
\usepackage{amsfonts}
\usepackage{amsthm}

\usepackage{url}
\usepackage{subfigure}
\usepackage{graphicx}
\usepackage[linesnumbered,ruled,vlined]{algorithm2e}
\usepackage{algpseudocode}
\usepackage{natbib}
\usepackage{wrapfig}
\usepackage{color}
\usepackage{epstopdf}
\usepackage[titletoc,toc,title]{appendix}
\usepackage{enumitem}
\usepackage{authblk}
\usepackage{epstopdf}
\usepackage{epsfig}
\usepackage{multicol}
\usepackage{multirow}
\usepackage{subfigure}
\usepackage{hyperref}
\usepackage{numprint}
\npthousandsep{,}\npthousandthpartsep{}

\renewcommand{\cite}{\citep}

\usepackage[skip=0.5\baselineskip]{caption}

\begin{document}


\title{Robust Spoken Language Understanding via Paraphrasing}

\author{Avik Ray \thanks{avik.r@samsung.com}}
\author{Yilin Shen \thanks{yilin.shen@samsung.com}}
\author{Hongxia Jin \thanks{hongxia.jin@samsung.com}}
\affil{{Samsung Research America} \authorcr
{Mountain View, CA, USA}}

\date{$17^{th}$ September $2018$}

\maketitle
\begin{abstract}
Learning intents and slot labels from user utterances is a fundamental step in all spoken language understanding (SLU) and dialog systems. State-of-the-art neural network based methods, after deployment, often suffer from performance degradation on encountering paraphrased utterances, and out-of-vocabulary words, rarely observed in their training set. We address this challenging problem by introducing a novel paraphrasing based SLU model which can be integrated with any existing SLU model in order to improve their overall performance. We propose two new paraphrase generators using RNN and sequence-to-sequence based neural networks, which are suitable for our application. Our experiments on existing benchmark and in house datasets demonstrate the robustness of our models to rare and complex paraphrased utterances, even under adversarial test distributions.    
\end{abstract}

%
%


\section{Introduction} \label{sec:intro}

Voice controlled personal agents (e.g. Alexa, Google Assistant, Bixby) are becoming popular due to their ability to understand a wide variety of user utterances, and perform different actions/tasks as requested by the user. Spoken language understanding (SLU) unit, or a semantic parser lie at its core which enables the agent to map a user utterance to the corresponding action desired by the user. Commercial semantic parsers represent the meaning of an utterance in terms of {\em intent} and {\em slot} labels, which can then be mapped to an action. {\em Intent detection} refers to the sub task of classifying an utterance into a semantic intent label, where as {\em slot tagging} is the sub task of providing a slot label to each word in the utterance. 

Traditional approaches treat intent detection as a semantic classification problem, and slot tagging as a sequence labeling problem. A wide variety of algorithms have been proposed e.g. SVMs \cite{HafTurWri:03}, hidden Markov models \cite{WangDengAce:05}, CRFs \cite{RayRic:07}, and more recently neural networks \cite{XuSar:13,MesDauYaoDengHakHe:15,HakTurCelChenGao:16,LiuLane:16}. State-of-the-art deep neural network models are trained jointly to solve the two tasks simultaneously using recurrent and sequence-to-sequence networks \cite{HakTurCelChenGao:16,LiuLane:16,KimLeeStratos:17,WangShenJin:18}. These models are trained end to end using labeled training data in the form of (utterance, intent label, slot labels) tuple. However, such datasets are expensive to collect, and are never exhaustive. As a result, after deployment, these data driven models suffer from poor accuracy on utterances which occur infrequently in their training data e.g. utterances with out-of-vocabulary words as well as various sentential paraphrases of the training utterances. The fundamental difficulty stems due to shortcoming of these models trained using likelihood maximization objective, that they do not generalize well to rare examples in training data. Unfortunately, this occur often in personal agent applications, since each individual user has their own personal vocabulary and paraphrase preferences.       

In this work, we try to tackle this problem by making the following important observation; often these infrequent and personalized user utterances have a paraphrased utterance which is more frequent in the training data. We try to answer the question; {\em instead of building a parser which perform well even for infrequent utterances, can we simply map such utterances to an utterance observed more frequently in the training data?} Subsequently, we can parse this more frequent utterance to understand meaning of the original utterance. Towards this end, we propose a new modular paraphrase driven parsing model, which can be integrated with any existing parser, to make it more robust to out-of-vocabulary and paraphrased utterances. In our proposed hybrid approach, we augment a parser with a paraphrase generator, which can be used to map an infrequent utterance to a more frequent paraphrased utterance. Traditional neural paraphrase generators, trained on large paraphrase corpus, however do not perform well in our setting with limited parser training data. Therefore, we further develop novel RNN and multi-task sequence-to-sequence based paraphrase generators, as well as techniques to build custom paraphrase datasets for their training. In our experiments, on both benchmark and custom in house datasets, we show that our hybrid paraphrase driven parsers can improve both accuracy and robustness of existing state-of-the-art and commercial parsers.     


\section{Problem and background} \label{sec:background}

\begin{figure}[hptb]
\begin{center}
\includegraphics[height =1in]{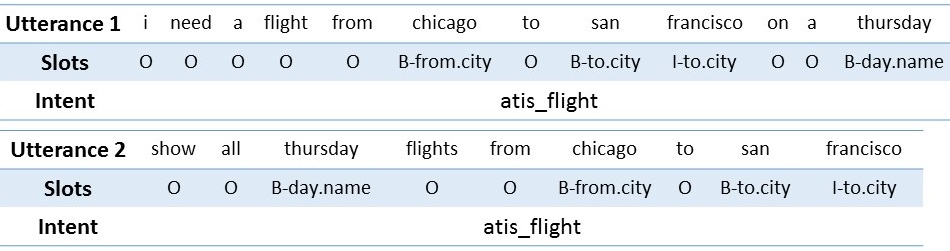} 
\caption{Examples of labeled utterances from our paraphrase dataset generated from ATIS training corpus.\label{fig:atis_examples} }
\end{center}
\end{figure}

In this section we formally define the intent classification and slot labeling problem, as well as discuss existing approaches. We are provided with a labeled training dataset $T=\{\mathbf{x}_i,\mathbf{y}_i,I_i\}_{i=1}^N,$ where $\mathbf{x}_i$ are the utterances with words in a vocabulary $\mathcal{V}_T,$ $\mathbf{y}_i$ represent the sequence of slot tags from a slot vocabulary $\mathcal{S},$ and $I_i \in \mathcal{I}$ represent the intent label of the utterance. A SLU unit consists of a parser $\mathcal{P}$ which can map an utterance $\mathbf{x}$ to its slot and intent labels $(\mathbf{y}, I).$ Figure \ref{fig:atis_examples} shows some example labeled utterances from benchmark ATIS dataset.

\noindent{\bf Recurrent and sequence-to-sequence models:} State-of-the-art and commercial neural network based parsers often use a single sequence-to-sequence and recurrent network to jointly infer the intent and slot labels \cite{HakTurCelChenGao:16,LiuLane:16,KimLeeStratos:17,WangShenJin:18}. Such encoder--decoder based deep neural networks for sequence learning have received considerable attention in the recent past due to its success in a variety of NLP tasks e.g. machine translation \cite{ChoBart:14,SutVinLe:14seq2seq}, parsing \cite{VinLuk:15,JiaLia:16,DongLap:16}, text generation \cite{RushChoWes:15}, paraphrasing \cite{PraHasLee:16} and so on.  Incorporating more encoder side information during decoding has been shown to further improve performance of these models \cite{BahChoBen:15}. A basic sequence-to-sequence neural network consists of an encoder $E$, and a decoder $D$, where each of them can be made up of multiple stacked recurrent units (e.g. LSTM). An input sequence $\mathbf{x}=(x_1,\hdots,x_n)$ is first encoded by repeatedly passing consecutive input symbols and previous hidden state $h_{t-1}$ through the encoder unit, $t\in [n].$ During decoding the decoder is first initialized with the final hidden state of the encoder, also called the context vector $c=h_n.$ Subsequently, the decoder hidden state $h_t'$ is updated at the decoder using the previous hidden state $h'_{t-1}$ and output symbol $y_{t-1},$ $t \in [m].$ The outputs are predicted using a softmax of the projected hidden decoder states as $p(y_t=y|y_{<t})=softmax(W_o h_t')\mathbf{1}_y,$ using a projection matrix $W_o.$ A sequence-to-sequence model is trained by maximizing the likelihood function:
\begin{equation}
p(y_1, .., y_m | x_1, .. , x_n ) = \Pi_{t=1}^m p(y_t|y_1 ,.., y_{t-1},c) \label{eq:seq_likelihood}
\end{equation}

\section{Our models} \label{sec:model}

Neural network parsers suffer from poor generalization on examples seen infrequently in their training data. In voice controlled personal agents this is of major concern since individual users often like to use their own personalized vocabulary and paraphrased utterances, which may not be present in the training data. Instead of adapting the parser directly to infrequent examples, we can choose to pre-process the original input to a more frequent example in the training data. This motivates our hybrid paraphrase driven parser for SLU discussed next.

The key idea behind our model is the following. Suppose there exists a base parser $\mathcal{P}_{base},$ trained using a dataset $T$ with vocabulary $\mathcal{V}_T.$ We augment this base parser with a paraphrase generator $\mathcal{P}_{para}$ trained on paraphrases from the training dataset $T,$ or unlabeled user log dataset. Now, when the base parser is unable to find the intent and slots of an infrequent utterance $\mathbf{x}$ with sufficient confidence, it chooses to retrieve a more frequent paraphrase of this utterance $\mathbf{x}'$ using the paraphrase generator $\mathcal{P}_{para}.$ The base parser then proceeds to infer the intent and slots from this paraphrased utterance $\mathbf{x}'.$ Since the paraphrase generator finds a frequent paraphrase $\mathbf{x}',$ the base parser is expected to achieve a higher parsing confidence on this new utterance. In essence the paraphrase generator acts as a translator between the user and $\mathcal{P}_{base}.$ 

\begin{figure}[hptb]
\begin{center}
\includegraphics[height =1.8in]{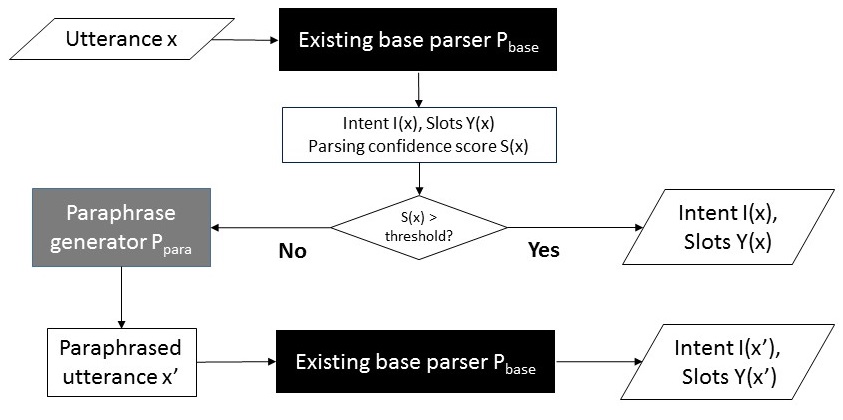}
\caption{Flowchart illustrating our paraphrase generator augmented robust parser for intent classification and slot tagging.\label{fig:para} }
\end{center}
\end{figure}

\noindent{\bf Algorithm:} Our paraphrase based parsing algorithm works as shown in Figure \ref{fig:para}. Suppose for each utterance $\mathbf{x},$ the base parser generates a confidence score $S(\mathbf{x})$ on the quality of the inferred intent and slots. Although we would like the paraphrase generator to paraphrase infrequent utterances, we do not want it to negatively effect the performance of the base parser $\mathcal{P}_{base}.$ Therefore, only the utterances with low parsing confidence $S(\mathbf{x}) < \tau$ are sent for paraphrasing. Computing confidence score of neural network output has been studied in various applications e.g. question answering \cite{GonLalKalMur:12}, semantic parsing \cite{DongQuiLap:18}. We compute a separate confidence score of the output intent label as the probability of the label from output softmax layer, $S_{intent}(\mathbf{x})=P(I=\ell),$ $\ell \in \mathcal{I}.$ Similarly, using the output probability of each of the slot tags, we compute an overall slot tagging score $S_{slot}(\mathbf{x}) = \exp \left(\frac{1}{m} \sum_{j}\log P(y_j=s_j) \right),$ $s_j \in \mathcal{S},$ using the normalized log likelihood of the tag sequence. The final score is computed as the minimum of these two scores $S(\mathbf{x}) = \min \{S_{intent}(\mathbf{x}), S_{slot}(\mathbf{x})\}.$   

We next describe the design of our paraphrase generator $\mathcal{P}_{para}.$ Many paraphrase generation techniques have been studied in literature \cite{KauBar:06,ZhaoWangTing:08,QuiBroDol:04,ZhaoNiuZhoLiuLi:08,ZhaoLanLiuLi:09}. However, most of these require additional labeled paraphrase training data which may not be always available. More importantly, these techniques do not guarantee that the generated paraphrase $\mathbf{x}'$ of $\mathbf{x}$ is a more frequent example which is well understood by the base parser. Therefore, we design two new paraphrase generation algorithms which are most suited for our parsing application. The first algorithm leverages an in--domain RNN language model to generate paraphrases using multiple word replacement. The second algorithm employs a neural paraphrase generation technique using a multi-task sequence-to-sequence model. 

\subsection{RNN language model based paraphrase generator}

RNN based language models have been widely used in automated speech recognition \cite{MikKarBurCerKhu,SunSchNey}. We use a similar language model as the main component in our first paraphrase generator. Note that, we require our paraphrases to be similar to training utterances of the base parser $P_{base}.$ Therefore we leverage the same training data, without any labels, to train our RNN language model. When the training dataset is small, this may not be sufficient to obtain a well trained language model. In such cases, we can leverage the large un-annotated user log data of a deployed personal agent, to train our language model. Such user log data is easily available in practical applications. We also train simultaneously two language models; $\mathcal{L}_f$ in the forward direction which predicts the probability of the $i^{th}$ word $w_i$ as $P(w_i| w_{i-1},...,w_{i-k})$; and $\mathcal{L}_b$ in the backward direction using a reversed corpus, predicting $P(w_i| w_{i+1},...,w_{i+k}),$ for a chosen $k$. Next we describe how these two language models are used for paraphrase generation.   

This model is motivated by the following observation. We use the term {\em context words} as words having the slot label ``O'' (which are non-informational), and {\em slot words} as the remaining informational words. For example in Figure \ref{fig:atis_examples}, words \{``chicago'', ``san francisco'', ``thursday''\} are slot words, and the remaining are context words. We observe that, often when the base parser $P_{base}$ fail to identify the correct slot labels, it can still identify the position of the slot words (but not their exact labels) with sufficient confidence. Since, context words play a major role in enabling identification of slot words, we would like to replace context words having low parser confidence with more frequent words, thereby generating a new paraphrase. This enables the slot words to be correctly labeled using this paraphrase. After $P_{base}$ identifies slot words in utterance $\mathbf{x},$ we assume the remaining words are context words, and find the average slot confidence $\overline{S}_C(\mathbf{x})$ over these context words $C.$ We generate a paraphrase template $T(\mathbf{x})=(u_i,\hdots, u_n)$ as follows; $u_i = x_i,$ if slot probability $P(y_i=s_i) > \overline{S}_C(\mathbf{x}),$ or if $x_i$ is a slot word, else we replace $u_i=\langle ? \rangle,$ a special blank token. We then run a modified beam search algorithm using the forward language model $\mathcal{L}_f$ over the template $T(\mathbf{x}),$ such that the beams are constrained to generate $u_i=x_i$ for non blank tokens, but are allowed to generate new words to replace the blank tokens $\langle ? \rangle$ in the template. However, these hard constraints tend to reduce the normal beam search quality of the RNN. To mitigate this, we also perform a similar reverse beam search using a reversed language model $\mathcal{L}_b.$ Finally, all generated beams are scored by both language models, and the one having the highest average score is output as paraphrase $\mathbf{x}'.$ As an example, a possible template $T(\mathbf{x})$ for utterance $1$ in Figure \ref{fig:atis_examples} is {\em ``$\langle ? \rangle$ $\langle ? \rangle$ a flight from chicago to san francisco on $\langle ? \rangle$ thursday''}; after beam search this may produce a paraphrase {\em ``show me a flight from chicago to san francisco on next thursday''}.   

\subsection{Multi-task neural paraphrase generator}

The paraphrases generated by RNN language model based generator can improve the slot identification performance of a parser (shown in Section \ref{sec:experiments}). However, the parser may still fail to correctly determine intent when the input utterance $\mathbf{x}$ is a structural paraphrase of some training utterance. Word replacement based paraphrase generators can never produce such structural variation. To tackle this issue our second paraphrase generator uses a neural multi-task sequence-to-sequence model. 

\begin{figure}[hptb]
\begin{center}
\includegraphics[height =1.8in]{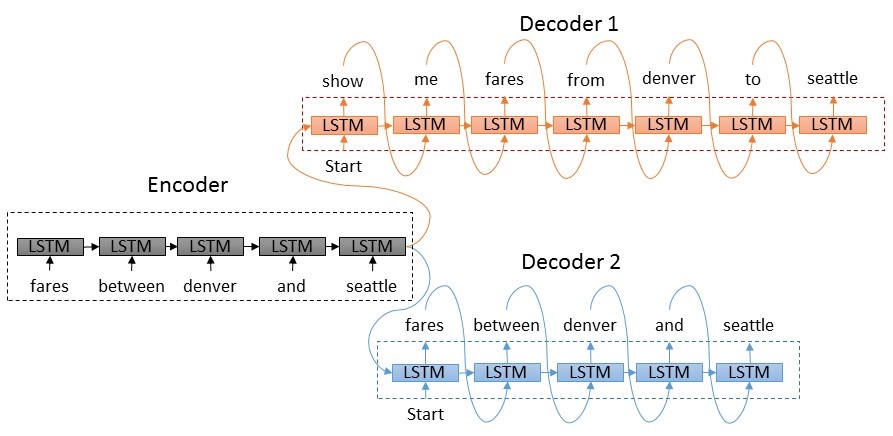}
\caption{Figure showing the architecture and training strategy of our multi-task sequence-to-sequence paraphrase generator. Additional attention structure from encoder to both decoders has been omitted in the diagram for clarity. \label{fig:multi_task} }
\end{center}
\end{figure}

Sequence-to-sequence based neural paraphrase generator has been proposed recently by Prakash et al. \cite{PraHasLeeDat:16}. However, we observe that the basic attention based sequence-to-sequence model do not perform well in our setting due to difficulty in paraphrasing utterances with rare slot words. Our paraphrase generator incorporates a single sequence encoder $E$, but two separate sequence decoders $D_1$ and $D_2$ as shown in Figure \ref{fig:multi_task}. During forward pass, both the decoders are initialized with the same encoder context vector $c,$ and then proceeds to decode the sequences independently. However, during training, we constraint the second decoder $D_2$ to generate the exact same input utterance $\mathbf{x},$ while the first decoder generates the paraphrase $\mathbf{x}'.$ Such additional autoencoder constraint has also been used in models for domain adaptation in order to obtain a better hidden representation vector of an utterance \cite{KimStraKim:17}. In our application, this better shared hidden representation encourages the correct reproduction of slot words even at the first decoder output. The model is trained using the joint multi-task objective function of the sum of the individual sequence loss functions at decoders $D_1, D_2.$ In addition, as a metric to determine the quality of the model during validation, we use a sum of BLEU score between input $\mathbf{x}$ and decoder $1$ output $\mathbf{x}',$ and the reconstruction accuracy of the input at decoder $2.$ Note that, although decoder 2 is trained as an autoencoder, during inference it may not always produce the same sequence as the input. We observe that decoder 2 output is also often a paraphrase of $\mathbf{x},$ having less structural variation. Therefore, we can use both the decoder outputs as paraphrases of $\mathbf{x},$ to be parsed by base parser $\mathcal{P}_{base}.$ \\

\noindent{\bf Paraphrase dataset generation:} In order to train our multi-task neural paraphrase generator, we generate a paraphrase dataset $T_{para}$ from the base parser training set $T$ as follows. First, we convert each utterance $\mathbf{x} \in T,$ to a {\em tagged utterance} where the slot words have been replaced by slot labels. For example, the tagged utterance corresponding to utterance 1 in Figure \ref{fig:atis_examples} is {\em``i need a flight from @from.city to @to.city on a @day.name''.} Now we observe that, tagged utterances having the same intent, and identical set of slot labels are paraphrases, since they are intended to convey the same meaning. This enables us to construct a tagged paraphrase dataset $T_{tagged}$ consisting of tuples of distinct tagged utterances $(\mathbf{z}, \mathbf{z}')$ which have the same intent and slot set. We then replace back the slot words from the parent utterance of $\mathbf{z}$ in both $\mathbf{z}$ and $\mathbf{z}',$ and vice versa. This generates the paraphrased dataset $T_{para}$ having tuples of paraphrases $(\mathbf{x}, \mathbf{x}').$ Figure \ref{fig:atis_examples} shows a paraphrase sample from this dataset. In addition, we also consider all training examples $\mathbf{x} \in T$ as identity paraphrases $(\mathbf{x}, \mathbf{x})$ and add them to $T_{para}.$ This prevents the paraphrase generator to perform poorly, when it encounters an utterance which did not have any other paraphrase in $T.$    


\section{Experiments} \label{sec:experiments}

In this section we describe our experimental results. We want to evaluate the intent and slot tagging accuracy gains using our paraphrase models compared to just a standalone parser. 

\begin{table}[ht]
\caption{Examples of complex utterances in our simulated ATIS log corpus.}\label{tab:ATIS_log}
\centering
\begin{tabular}{|c|p{5in}|}
\hline
{\bf Intent} & {\bf Utterance} \\\hline
atis flight & show me trip that leaves tuesday on american airline going from baltimore leaving early night arriving in pittsburgh  \\ \hline
 atis airfare & give me the fares with continental leaving from long beach for flights one way with first class arriving in tacoma \\ \hline \hline
\end{tabular} 
\end{table}

\noindent{\bf Datasets:} For evaluation we use the benchmark ATIS dataset \cite{AtisPilotHemp90atis}, which is popularly used for evaluating parsers for spoken language understanding. The ATIS dataset contain \numprint{5871} utterances related to airline reservation with \numprint{4978} training and \numprint{893} test utterances. Overall it contains \numprint{17} intent labels and \numprint{79} slot labels. Example utterances from this dataset is shown in Figure \ref{fig:atis_examples}. In order to show the accuracy gains for various sizes of training set, we further sample training sets of such sizes, but test parser performance on the full ATIS test set. 

\begin{table}[ht]
\caption{Comparison of $10$ fold average test intent accuracy percentage of all models on ATIS corpus with increasing size of training set, using the attention BiRNN \cite{LiuLane:16} as the base parser.}\label{tab:atis_accuracy_birnn}
\centering
\setlength\tabcolsep{4pt}
\begin{tabular}{|l|c|c|c|c|c|}
\hline
{\bf Parsers} & \multicolumn{5}{c|}{\textbf{Training dataset size}} \\\hline
 & $\mathbf{500}$ & $\mathbf{1500}$ & $\mathbf{2500}$ & $\mathbf{3500}$ & $\mathbf{4500}$ \\ \hline
BiRNN (Liu and Lane) & $87.95$ & $93.06$ & $94.26$ & $96.08$ & $96.47$ \\ \hline
Seq-to-seq paraphrase + BiRNN & $88.16$ & $93.10$ & $94.54$ & $\mathbf{96.26}$ & $\mathbf{96.58}$ \\ \hline
RNN paraphrase + BiRNN & $\mathbf{88.63}$ & $\mathbf{93.39}$ & $\mathbf{94.55}$ & $96.20$ & $96.52$ \\ \hline \hline
\end{tabular} 
\end{table}

After deployment of an intelligent personal agent, often the distribution of the observed utterances turn out to be significantly different from those used in training. This is because each individual users have their own preferred choice of paraphrase and vocabulary, and this can change over time \cite{KimStraKim:17}. In order to test the robustness of our models in such adversarial scenario, we generate a simulated {\em  ATIS log} dataset as follows. Starting with the original ATIS dataset, we use data recombination techniques similar to \cite{JiaLia:16}, to generate a variety of long and complex utterances. Then, human linguistic experts prune any incorrect utterance. We train all models on the original ATIS training set, then we test their performance on a set of \numprint{1000} ATIS log utterances for testing. Example utterances from our dataset are shown in Table \ref{tab:ATIS_log}.

\noindent{\bf Baselines and parameters:} We use two baseline parsers for evaluation. First, we use the state-of-the-art {\em Attention BiRNN} based neural network parser by Liu and Lane \cite{LiuLane:16}. As a second baseline parser we use the open source {\em RASA} parser \cite{RASA}, in order to demonstrate the applicability in commercial agents and dialog systems. We augment both these parsers with our paraphrase generation models and compare their performance with the former. For attention BiRNN, we use the Tensorflow implementation made available by Liu et al. with its default parameters. We also use the RASA parser with its default settings. Our neural models were implemented in Tensorflow. The confidence threshold $\tau$ in our paraphrase models were set to $0.8.$ For RASA we were unable to use RNN based paraphrase model, since it does not return the slot tagging probabilities required for paraphrase template construction. In paraphrase models, we generate two best paraphrases using the paraphrase generators, and perform a simple majority voting to predict the final intent and slot labels.

\subsection{Results}

First we compare the performance of different models on the benchmark ATIS dataset. In Table \ref{tab:atis_accuracy_birnn} we compare the \numprint{10} fold average intent detection accuracy of our models when combined with the attention BiRNN baseline model. We observe that both our models improve the accuracy of the baseline parser. Further, the accuracy gain is higher when the training set size is small. In Table \ref{tab:atis_f1_birnn} we compare their corresponding average slot tagging F1 scores. The RNN paraphrase model is observed to improve F1 score of the baseline model, while the sequence-to-sequence model do not. This is expected, since the RNN paraphrase model only replaces the low confidence context words with more frequent words, which enables the base parser to better identify the slot labels. In contrast, the sequence-to-sequence paraphrase model may alter the sentence structure and slot words in its paraphrases, hence doesn't always improve slot tagging F1 score. We further observe that it often improves the recall, but not its precision.

\begin{table}[ht]
\caption{Comparison of $10$ fold average slot tagging F1 score percentage of all models on ATIS corpus with increasing size of training set, using the attention BiRNN \cite{LiuLane:16} as the base parser.}\label{tab:atis_f1_birnn}
\centering
\setlength\tabcolsep{4pt}
\begin{tabular}{|l|c|c|c|c|c|}
\hline
{\bf Parsers} & \multicolumn{5}{c|}{\textbf{Training dataset size}} \\\hline
 & $\mathbf{500}$ & $\mathbf{1500}$ & $\mathbf{2500}$ & $\mathbf{3500}$ & $\mathbf{4500}$ \\ \hline
BiRNN (Liu and Lane) & $79.96$ & $88.57$ & $90.83$ & $\mathbf{91.33}$ & $92.02$ \\ \hline
Seq-to-seq paraphrase + BiRNN & $79.83$ & $88.44$ & $90.76$ & $91.28$ & $91.98$ \\ \hline
RNN paraphrase + BiRNN & $\mathbf{80.01}$ & $\mathbf{88.62}$ & $\mathbf{90.84}$ & $91.29$ & $\mathbf{92.02}$ \\ \hline \hline
\end{tabular} 
\end{table}

Next, we compare the performance of the sequence-to-sequence paraphrase model, when used with RASA as the base parser. RASA uses a kernel SVM classifier along with feature selection. Hence, in general it has a worse performance than attention BiRNN parser. However, it has the advantage of fast training time. Table \ref{tab:atis_accuracy_rasa} compares both the intent and slot tagging performance. Once again, we observe that the paraphrase model is able to achieve high gains in intent accuracy over the baseline RASA parser. The slot tagging performance do not improve with this model as previously observed with attention BiRNN parser. As mentioned before, due to the lack of slot tagging confidence scores in RASA, we are unable to use our RNN based paraphrase model with RASA.

\begin{table}[ht]
\caption{Comparison of $10$ fold average test intent accuracy and slot tagging F1 score percentages of the sequence-to-sequence model on ATIS corpus with increasing size of training set, using RASA \cite{RASA} as the base parser.}\label{tab:atis_accuracy_rasa}
\centering
\setlength\tabcolsep{4pt}
\begin{tabular}{|l|c|c|c|c|c|}
\hline
{\bf Parsers} & {\bf Metric} & \multicolumn{4}{c|}{\textbf{Training dataset size}} \\\hline
 & & $\mathbf{500}$ & $\mathbf{1500}$ & $\mathbf{2500}$ & $\mathbf{3500}$ \\ \hline
RASA & Accuracy & $83.70$ & $86.81$ & $88.33$ & $88.44$ \\ \hline
Seq-to-seq paraphrase + RASA & Accuracy & $\mathbf{84.64}$ & $\mathbf{89.32}$ & $\mathbf{90.06}$ & $\mathbf{91.62}$ \\ \hline \hline
RASA & F1 & $75.39$ & $81.34$ & $83.41$ & $84.71$ \\ \hline
Seq-to-seq paraphrase + RASA & F1 & $75.25$ & $81.32$ & $83.22$ & $84.55$ \\ \hline \hline
\end{tabular} 
\end{table}

Finally, to validate the robustness of our models in an adversarial post deployment scenario, we test the performance of our models on the simulated ATIS log corpus. Table \ref{tab:para_atis_accuracy_all} reports the intent classification accuracy of all our models. Due to a distribution mismatch with the ATIS training data, all models perform worse in this dataset, as expected. However, we still observe that, irrespective of the base parser used, our paraphrase models achieve an improved intent detection accuracy.

\begin{table}[ht]
\caption{Comparison of $10$ fold average test intent accuracy percentage of all our models on simulated ATIS log corpus with increasing size of training set, using both attention BiRNN \cite{LiuLane:16} and RASA \cite{RASA} as the base parser. The models are trained on original ATIS dataset but tested on ATIS log corpus.}\label{tab:para_atis_accuracy_all}
\centering
\setlength\tabcolsep{4pt}
\begin{tabular}{|l|c|c|c|c|}
\hline
{\bf Parsers} & \multicolumn{4}{c|}{\textbf{Training dataset size}} \\\hline
 & $\mathbf{500}$ & $\mathbf{1500}$ & $\mathbf{2500}$ & $\mathbf{3500}$ \\ \hline
BiRNN (Liu and Lane) & $80.49$ & $82.22$ & $82.71$ & $82.65$ \\ \hline
Seq-to-seq paraphrase + BiRNN & $82.41$ & $82.51$ & $83.54$ & $82.87$ \\ \hline
RNN paraphrase + BiRNN & $\mathbf{83.25}$ & $\mathbf{83.31}$ & $\mathbf{84.24}$ & $\mathbf{83.22}$  \\ \hline \hline
RASA  &  $77.60$ & $83.16$ & $83.56$ & $84.26$ \\ \hline
Seq-to-seq paraphrase + RASA & $\mathbf{79.66}$ & $\mathbf{85.14}$ & $\mathbf{86.08}$ & $\mathbf{84.52}$ \\ \hline \hline
\end{tabular} 
\end{table}



\section{Conclusion} \label{sec:conclusion}

Commercial parsers trained using data driven approaches, often have poor performance after deployment, when it encounters a variety of complex paraphrases and out-of-vocabulary words that were unseen or infrequent in its training data. In this paper, we propose a novel paraphrase driven parsing approach, where during parsing such complex paraphrases are first converted to a more familiar utterance using a paraphrase generator. We propose two new paraphrase generation techniques suitable to use in our application. Our experimental results validate that, irrespective of the base parser being used, or the test data distribution being observed, our combined models are able to greatly improve the performance of the standalone base parser. 


%

\bibliographystyle{abbrvnat}
\bibliography{intent_parse}

%
%

\end{document}